\documentclass[letterpaper]{article} 
\usepackage{aaai2026}
\usepackage{times}  
\usepackage{helvet}  
\usepackage{courier}  
\usepackage[hyphens]{url}  
\usepackage{natbib}  
\usepackage{caption} 
\usepackage{algorithm}
\usepackage{algorithmic}
\usepackage{url}
\usepackage{newfloat}
\usepackage{listings}
\DeclareCaptionStyle{ruled}{labelfont=normalfont,labelsep=colon,strut=off} 
\floatstyle{ruled}
\newfloat{listing}{tb}{lst}{}
\floatname{listing}{Listing}

\usepackage{microtype}
\usepackage{subfigure}
\usepackage{booktabs} 
\usepackage[abbreviations]{glossaries-extra}
\usepackage{multirow}
\usepackage{xspace}
\usepackage{pifont}

\usepackage{amsmath}
\usepackage{amssymb}
\usepackage{mathtools}
\usepackage{amsthm}
\usepackage{array} 

\newcommand{\refFig}[1]{Fig.~\ref{fig:#1}}

\newcommand{\refTbl}[1]{Table.~\ref{tbl:#1}}

\newabbreviation{ViCLAS}{ViCLAS}{Violent Crime Linkage Analysis System}
\newabbreviation{SCAS}{SCAS}{Serious Crime Analysis Section}
\newabbreviation{NCA}{NCA}{National Crime Agency}
\newabbreviation{CL}{CL}{Crime Linkage}
\newabbreviation{MO}{MO}{Modus Operandi}
\newabbreviation{PCA}{PCA}{Principal Component Analysis}
\newabbreviation{AUC}{AUC}{Area Under the ROC Curve}
\newabbreviation{TPFP}{TPFP}{True Positive Rate at Fixed False Positive Rate}
\newabbreviation{AUPRC}{AUPRC}{Area Under the Precision-Recall Curve}
\newabbreviation{MLP}{MLP}{Multilayer Perception}
\newabbreviation{CNN}{CNN}{Convolutional Neural Network}
\newabbreviation{RQ}{RQ}{Research Question}
\newabbreviation{Std}{Std}{Standard Deviation}
\newabbreviation{ML}{ML}{Machine Learning}
\newabbreviation{CV}{CV}{Cross Validation}

\global\long\def\CV{\gls{CV}\xspace}
\global\long\def\ML{\gls{ML}\xspace}
\global\long\def\Std{\gls{Std}\xspace}
\global\long\def\RQ{\gls{RQ}\xspace}
\global\long\def\CNN{\gls{CNN}\xspace}
\global\long\def\MLP{\gls{MLP}\xspace}
\global\long\def\AUC{\gls{AUC}\xspace}
\global\long\def\TPFP{\gls{TPFP}\xspace}
\global\long\def\AUPRC{\gls{AUPRC}\xspace}
\global\long\def\ViCLAS{\gls{ViCLAS}\xspace}
\global\long\def\SCAS{\gls{SCAS}\xspace}
\global\long\def\NCA{\gls{NCA}\xspace}
\global\long\def\CL{\gls{CL}\xspace}
\global\long\def\MO{\gls{MO}\xspace}
\global\long\def\PCA{\gls{PCA}\xspace}

\theoremstyle{plain}

\theoremstyle{definition}

\theoremstyle{remark}

\usepackage[textsize=tiny]{todonotes}

\title{Enhancing Binary Encoded Crime Linkage Analysis Using Siamese Network}

\author{
    Yicheng Zhan\textsuperscript{\rm 1},
    Fahim Ahmed\textsuperscript{\rm 1},
    Amy Burrell\textsuperscript{\rm 2},
    Matthew Tonkin\textsuperscript{\rm 3},
    Sarah Galambos\textsuperscript{\rm 4},
    Jessica Woodhams\textsuperscript{\rm 2},
    Dalal Alrajeh\textsuperscript{\rm 1}
}

\affiliations{
    \textsuperscript{\rm 1}Department of Informatics, Imperial College London, London, UK\\
    \textsuperscript{\rm 2}University of Birmingham, Birmingham, UK\\
    \textsuperscript{\rm 3}University of Leicester, Leicester, UK\\
    \textsuperscript{\rm 4}UK National Crime Agency, Serious Crime Analysis Section, UK\\
    yz10621@imperial.ac.uk, dalal.alrajeh04@imperial.ac.uk, j.woodhams@bham.ac.uk
}

\begin{document}

\maketitle

\begin{abstract}
  Effective crime linkage analysis is crucial for identifying serial offenders and enhancing public safety. 
 To address the limitations of traditional crime linkage methods when handling high-dimensional, sparse, and heterogeneous data, this paper proposes a Siamese Autoencoder framework to learn meaningful latent representations and uncover correlations in highly complex data. 
Using a dataset from the Violent Crime Linkage Analysis System—a database maintained by the Serious Crime Analysis Section of the UK’s National Crime Agency—our approach mitigates signal dilution in high-dimensional sparse data through decoder-stage integration of geographic-temporal features. This integration amplifies learned behavioral representations rather than allowing them to be overwhelmed at the input stage, leading to consistent improvements over baseline methods across multiple metrics.
  We further examine how different data reduction strategies based on domain-expert can impact model performance, offering practical insights into preprocessing for crime linkage. 
 Our solution shows that advanced machine learning approaches can enhance linkage accuracy, improving AUC by up to 9\% over traditional methods and providing insights to support human decision-making in crime investigation.
\end{abstract}

\section{Introduction}
\label{sec:intro}

\CL is integral to modern law enforcement, aiming to identify 
various types of serial offences (e.g., serial sexual assault, burglary, or robbery) and can also help link crimes to a known offender. The process focuses on the offender's \MO, a specific combination of behaviours that differentiate one offender from another, during the commission of the crimes to predict that the same individual is responsible for multiple incidents. 
Accurate \CL assists in prioritizing investigative resources and strengthening public safety measures. 
Real-world crime databases, however, often contain complex data landscapes marked by high dimensionality, sparsity, and geographic-temporal heterogeneity \cite{grubin2001linking}. While traditional methods have demonstrated efficiency in controlled scenarios, their effectiveness diminishes when deployed on datasets exhibiting these real-world complexities.

This performance gap becomes apparent in large \CL systems. In this paper, we examine our method using a dataset derived from the \ViCLAS, a database management system, used by the \SCAS of the UK’s \NCA. 
\ViCLAS is the largest crime dataset analyzed in published research to date. 

Moreover, conventional \CL research on serious sexual offending primarily focused on patterns in the \MO alone, with limited consideration of geographic-temporal information of offences.
While some \ML approaches explored geographic-temporal patterns in specific contexts (e.g., standalone analyses of burglary distance metrics and temporal intervals~\cite{solomon2020crime}), they mainly focused on smaller-scale, geographically-confined datasets.

We propose a Siamese Autoencoder framework that jointly learns latent representations from complex, high-dimensional binary-encoded data. Since geographic-temporal features become statistically insignificant when concatenated with behavioral features, we introduce decoder-stage integration to preserves their discriminative characteristics by modulating learned embeddings rather than competing during feature extraction. This choice yields consistent improvements of 0.86-3.29\% AUC across network variants (\refTbl{architecture_comparison}).
To address the high dimensionality of \ViCLAS data, 
we further explore the impact of domain-specific, expert-defined data reduction strategies for dimensionality reduction and evaluate their effectiveness. 
This paper aims to examine how \ML techniques can support crime linkage analysis, aiming to accelerate and help investigative decision-making. 
Our experiments on real-world data show that the proposed approach outperforms both traditional methods and Naive Siamese baselines. The sanitized code is available at: \url{https://github.com/AlberTgarY/CrimeLinkageSiamese}.

\begin{itemize}
    \item A novel application of Siamese Autoencoder to crime linkage analysis, evaluated on \ViCLAS, the largest real-world dataset on serious sexual offences analyzed in published research.
    \item An exploration of data reduction strategies that highlights effective preprocessing techniques based on domain-expert for sparse, high-dimensional datasets.
    \item Key insights into the application of \ML to crime data, including the impact of dataset properties and the sensitivity of network structures.
\end{itemize}

\section{Related Work}
\subsection{Crime Linkage}
Crime linkage identifies offence series and supports the detection and apprehension of prolific offenders~\cite{burrell2020behavioural}, enabling resource sharing and preventing redundant efforts~\cite{grubin2001linking}. While physical evidence (e.g., fingerprints, DNA) can directly link crimes, behavioural indicators—victim approach, violence level, weapon use—offer alternative linkage when forensic traces are absent~\cite{bennell2005between, tonkin2017using}. Behavioural Crime Linkage (BCL) rests on two theoretical pillars: behavioural consistency (similar offender actions across offences) and behavioural distinctiveness (unique patterns differentiating offenders)~\cite{woodhams_bennell_2014}. Empirical support for these assumptions is extensive~\cite{burrell2024methods}. In practice, practitioners employ both individual behaviours and thematic clusters representing shared functions~\cite{alison2011pragmatic}. Research has explored various multivariate linkage techniques, including taxonomic similarity metrics~\cite{woodhams2007marine}, discriminant function analysis and multidimensional scaling for theme derivation~\cite{winter2013comparing}, and other advanced methods~\cite{burrell2024methods}.

\subsection{Data-Driven Approaches in Crime Linkage}
\label{sec:data_driven}

Early crime linkage approaches relied on statistical models such as logistic regression and decision trees using predefined modus operandi (MO) features to capture behavioural consistencies~\cite{bennell2005between,melnyk2011another}. While effective for controlled scenarios, these models struggle with non-linear dependencies intrinsic to crime data. Recent advances leverage machine learning to process large datasets and uncover complex patterns~\cite{bennell2014linking}. Li and Qi~\cite{li2019approach} demonstrated enhanced serial-crime detection by combining natural language processing with dynamic time warping on crime narratives. However, challenges remain for sexual-offence linkage, where geographically dispersed, small samples often lead to under-exploited spatial and temporal features~\cite{woodhams2008incorporating,grubin2001linking}.

\subsection{Machine Learning Techniques in Crime Linkage}
\label{sec:machine_learning}
With the advancement of artificial intelligence, \ML methods have also been applied to crime linkage tasks. 
\cite{forecast3040046} examined various deep learning architectures for crime classification and prediction, showing the robustness of learned methods outperforming traditional methods when working with 2D incident images as input. 
Similarly, \cite{utsha2024deep} also reviewed deep learning-based crime prediction models and highlighted the effectiveness of specific architectures in dealing with sparse crime data. 
Furthermore, \cite{hotspot2023} explored the application of spatio-temporal neural networks for predicting crime hotspots, seeking optimal configurations for crime data analysis.
The recent study by \cite{burrell2024methods} presents a comprehensive review of the application of \ML and the other approaches in \CL.
One notable approach in the learned domain \CL is the use of Siamese Autoencoders. 
These networks optimize the similarity between linked crimes while maximizing the differences between unlinked ones, 
rather than relying on predefined similarity metrics.
\cite{solomon2020crime} applied Siamese neural networks to crime data, using embeddings derived from textual features combined with spatial-temporal information to predict linkages between burglary cases.
In this work, we extend the application of Siamese architectures by introducing a novel geographic-temporal integration approach designed to process high-dimensional binary-encoded features.

\section{Dataset: \ViCLAS}

We utilize data from  \ViCLAS, a comprehensive database maintained by the \SCAS of the \NCA in the United Kingdom. 
\ViCLAS is designed to capture detailed information about violent and sexual offences (weapon, victim, scene, vehicle, and other variables that are related to the actual
offence), providing a rich source of data for crime linkage analysis~\cite{Viclas2022MarkLaw}.
The original \ViCLAS dataset remains categorical in nature, with a few binary representations.
For research, the dataset is reformatted in binary, with 1 indicating an observed attribute and 0 for unobserved or unrecorded ones.\footnote{Access to the data used in this research was granted through requests R123, R128, R182a, and R182b submitted to \SCAS. Due to strict confidentiality and data-sharing agreements with the Agency, the data cannot be shared publicly. Access requests for research must be submitted directly to the UK's \NCA.} 
Our study employs two variants derived from this database:

\subsection{Single Victim-Offender-Scene Series}
Serving as a proof of concept for our approach, we first analyzed a focused subset of our main dataset, which was recorded on January 6, 2014. 
The dataset consists of a collection of series involving offenders convicted of multiple offences. Each offence was executed by a single offender against a single victim at a single scene. The simplification facilitates the attribution of behaviour exhibited in an offence to the offender. 
This initial dataset consists of 1,482 cases distributed across 493 series and does not contain geographic-temporal data. 
Each case is identified by a unique ViCLAS reference (ID), which serves as a linking key of the same offence across various data sheets.

\subsection{Multiple Victim-Offender-Scene Series}

For the main dataset, we expanded our analysis beyond the initial Single Victim-Offender-Scene Series to our main dataset spanning January 1990 to November 2021, comprising 22,282 offences across 446 features. The dataset includes both solved cases (where sufficient evidence links the offence to a known offender) and unsolved cases.
Unlike the initial dataset, the incident may involve multiple offenders against multiple victims, where the offence occurs across multiple scenes. There are no means for directly attributing which offensive behaviour was performed by which offender, against which victim, and at what scene. This increases the complexity of the dataset and the problem of identifying patterns.
Of these, 12,625 cases were categorized as solved, and 11,970 were retained for analysis after applying data validation steps to ensure consistency and completeness.
The final analysis focuses on 446 of the 449 features.
For the purpose of exploring data reduction strategies, we further adopted information about the type of features in the dataset,  i.e., behavioural or contextual, following the process in \cite{Viclas2022MarkLaw}.
Behavioural features are those that capture behaviours exhibited by the offender (e.g., weapon use, approach method, verbal threats). Contextual features describe the context in which the offence took place (e.g., location type, time of day, victim characteristics).
As such, the dataset encompassed 177 behavioural features and 158 contextual features, with 11 features classified as both behavioural and contextual.

\section{Methodology}
Effective \CL remains challenging due to the high-dimensional and sparse nature of crime data~\cite{CHI201788}. Traditional methods impose limiting assumptions: logistic regression assumes linear feature relationships while decision trees enforce rigid hierarchical structures, both inadequate for capturing non-linear criminal behavior associations. Meaningful patterns emerge from feature combinations rather than direct matching, which is particularly challenging given sparse binary encodings are dominated by zero values.

\begin{figure}[ht!]
  \centering
  \includegraphics[width=0.45\textwidth]{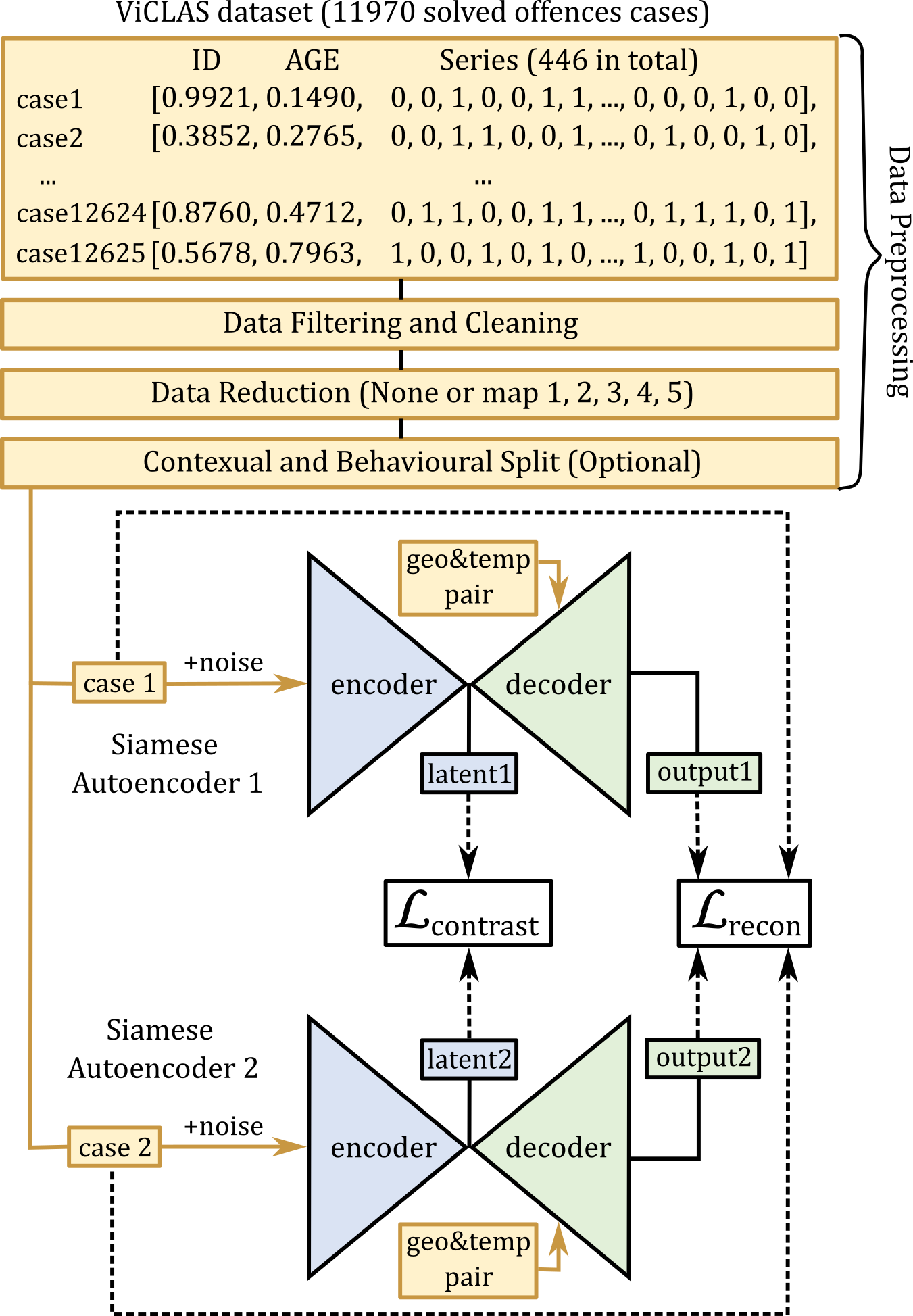}
  \caption{The overview of the network training pipeline. 
  Case data from the \ViCLAS dataset undergoes data filtering\&clearning and data reduction before being processed by the Siamese Autoencoder.
  The framework handles pairwise cases with added noise and incorporates geographic-temporal data at the decoder stage.
  Here geo\&temp refers to geographic-temporal data.
}
\label{fig:training_pipeline}
\end{figure}

We present a novel Siamese Autoencoder framework that discovers latent representations in \ViCLAS dataset and incorporates geographic-temporal data for \CL analysis. As shown in \refFig{training_pipeline}, 
our approach comprises three components: \textbf{(Network Structure)} extracts compact embeddings from sparse, binary-encoded data, with geographic-temporal data fused at the decoder stage; \
\textbf{(Loss Function)} employs contrastive and reconstruction losses to cluster crimes by the same offender while separating unlinked cases; 
\textbf{(Model Inference On Unsolved Cases)} transforms latent code distances into bounded probability scores for case comparison. We begin by describing the problem definition below.

\subsection{Problem Definition}
Given a set of $N$ binary-encoded criminal incidents, each described by a high-dimensional feature vector (e.g., behavioural and contextual features) and continuous geographic-temporal indicators (e.g., distance and time interval), the goal of \CL is to determine whether any two or more incidents originate from the same offender. Formally, let $\mathbf{x}_i \in \{0,1\}^{M}$ be the binary-encoded features of the $i$-th incident where $M=446$ represents the original feature dimensionality before reduction, and let $g(\mathbf{x}_i,\mathbf{x}_j) \in \mathbb{R}^2$ capture the continuous geographic-temporal data (specifically, the spatial distance and temporal interval) between incidents $i$ and $j$. We seek a function $f(\mathbf{x}_i, \mathbf{x}_j, g(\mathbf{x}_i,\mathbf{x}_j)) \rightarrow [0,1]$ that outputs a probability score reflecting whether these two incidents are linked (i.e., committed by the same offender), where higher scores indicate greater probability of linkage.

\subsection{Network Structure}

\begin{figure}[ht!]
  \centering
  \includegraphics[width=0.25\textwidth]{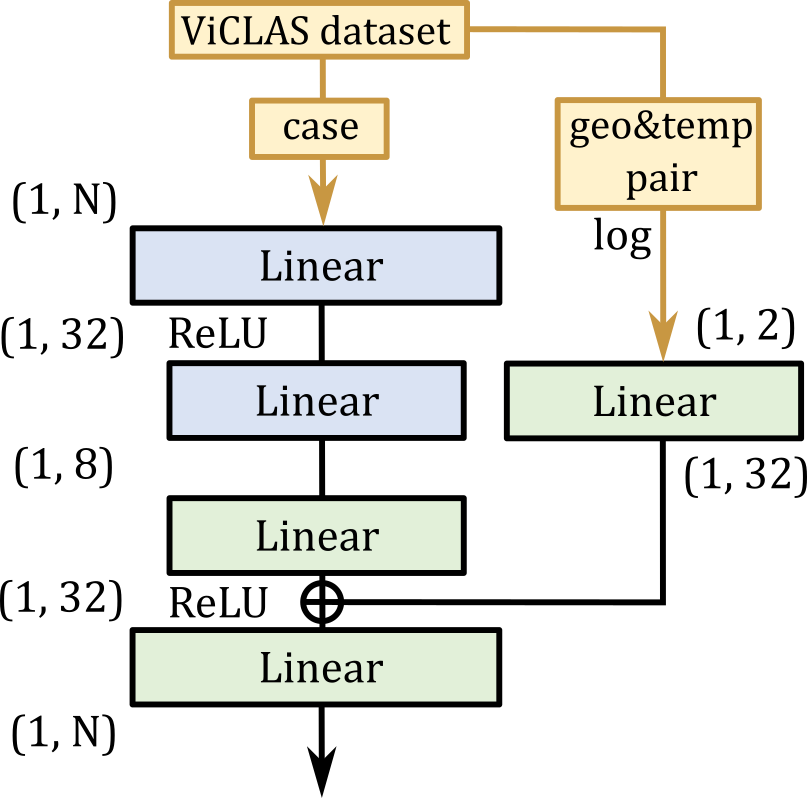}
  \caption{The architecture of our Siamese Autoencoder. 
  The encoder reduces the input dimensionality to 8, while the decoder reconstructs the latent code back to the original input shape.
  Here geo\&temp refers to geographic-temporal data.
}
\label{fig:siamese}
\end{figure}

\textbf{Architectural Overview.} Figure~\ref{fig:siamese} shows our Siamese Autoencoder with two identical sub-networks processing crime pairs in parallel. Each sub-network uses an encoder-decoder structure optimized for binary behavioral data. The encoder comprises two linear layers with ReLU activations (446→128→8), compressing input features to an 8-dimensional latent representation. The decoder mirrors this architecture (8→128→446), reconstructing the original feature space.
The decoder mirrors this structure, reconstructing original features. Geographic-temporal data integration occurs between decoder layers. Logarithmically transformed spatial-temporal features pass through a linear layer and combine additively with the first decoder layer output, preserving geographic-temporal influence. 
Our proposed network consists of 21,740 parameters, compared to the 22,981 parameters of the Naive Siamese network baseline. Despite having comparable parameters, our approach consistently demonstrates higher performance across metrics.

\label{sec:structure}
\textbf{Motivation.}
Our architecture extends \cite{solomon2020crime} by incorporating reconstruction constraints tailored for the \ViCLAS domain, targeting three key challenges:
(1) \emph{Feature Dimensionality}: Solomon et al. used only 40-dimensional TF-IDF vectors, whereas our model handles 217–446 binary-encoded dimensions, requiring compact representation learning from high-dimensional, sparse data. 
(2) \emph{Structurally-Informed Contrastive Learning}: We integrate reconstruction loss to retain structural information in latent representations, enhancing discriminative capability via contrastive learning.
(3) \emph{Geographic-Temporal Integration}: Instead of concatenating geographic-temporal features at the input (where their impact is minimal), we fuse them at the decoder stage for stronger signal amplification. 
For more information about the motivation of our architecture, please refer to the Supplementary Material.

\textbf{Decoder-Stage Integration Rationale.} Integrating geographic-temporal data at the decoder stage rather than at the input level addresses two issues:
(1) \emph{Signal Dilution}: When concatenated at the input layer, the 2-dimensional geographic-temporal data become statistically insignificant ($<1\%$) against 217-446 behavioral dimensions, limiting their discriminative potential. Decoder-stage integration enables explicit modulation of latent codes after behavioral abstraction, amplifying pairwise geographic-temporal signals where they provide maximum discriminative value.
(2) \emph{Pairwise Semantics}: Geo-temporal data inherently reflect pairwise relationships (comparisons between crimes). Incorporating this data after encoding stage allows latent behavioral representations to remain individually consistent within encoder, while the decoder adjusts for these pairwise geo-temporal relationships, mirroring real investigative practice. \refFig{training_pipeline} illustrates our complete training pipeline, where paired cases undergo data preprocessing, reduction, and simultaneous processing through twin networks to generate distance-based linkage predictions.

\subsection{Loss Function}
\label{sec:loss}
Our model jointly optimizes contrastive and reconstruction losses, defined as 
$\mathcal{L} = \alpha \mathcal{L}_{\text{contrast}} + \beta \mathcal{L}_{\text{recon}}$, 
with $\alpha = 1.0$ and $\beta = 0.2$. 
As identifying linked cases is the primary objective, the contrastive term $\mathcal{L}_{\text{contrast}}$ is given higher weight to maintain interpretability.

\paragraph{Contrastive Loss.}
$\mathcal{L}_{\text{contrast}}$ clusters cases by the same offender while separating dissimilar ones using a hybrid Euclidean-Manhattan distance metric~\cite{TONKIN201719}:
\begin{equation}
d(x_1, x_2) = \sqrt{\sum_{i=1}^{n}(x_{1i} - x_{2i})^2} + \sum_{i=1}^{n}|x_{1i} - x_{2i}|,
\end{equation}
where $x_1$ and $x_2$ denote latent representations. The contrastive objective is defined as
\begin{equation}
\mathcal{L}_{\text{contrast}} = \alpha \cdot \mathbb{E}[y \cdot d^2 + (1-y) \cdot \max(m-d, 0)^2],
\end{equation}
where $y$ indicates case linkage (1=linked, 0=unlinked), $m=5$ is the margin parameter (empirically determined from values 1-10 for optimal clustering-separation balance), and $\alpha=1$ is the scaling factor\cite{liu2023single, ghojogh2020fisher}.

\paragraph{Reconstruction Loss.}
$\mathcal{L}_{\text{recon}}$ ensures latent representations retain sufficient information for input reconstruction using cosine similarity:
\begin{equation}
\mathcal{L}_{\text{recon}} = \mathbb{E}\left[\frac{v_1^\top\hat{v_1}}{\|v_1\|\|\hat{v_1}\|} + \frac{v_2^\top\hat{v_2}}{\|v_2\|\|\hat{v_2}\|}\right],
\end{equation}
where $v_i$ and $\hat{v_i}$ are original and reconstructed feature vectors, $\|\cdot\|$ denotes L2 norm, and $\top$ represents transpose.

\begin{figure}[ht!]
  \centering
  \includegraphics[width=0.47\textwidth]{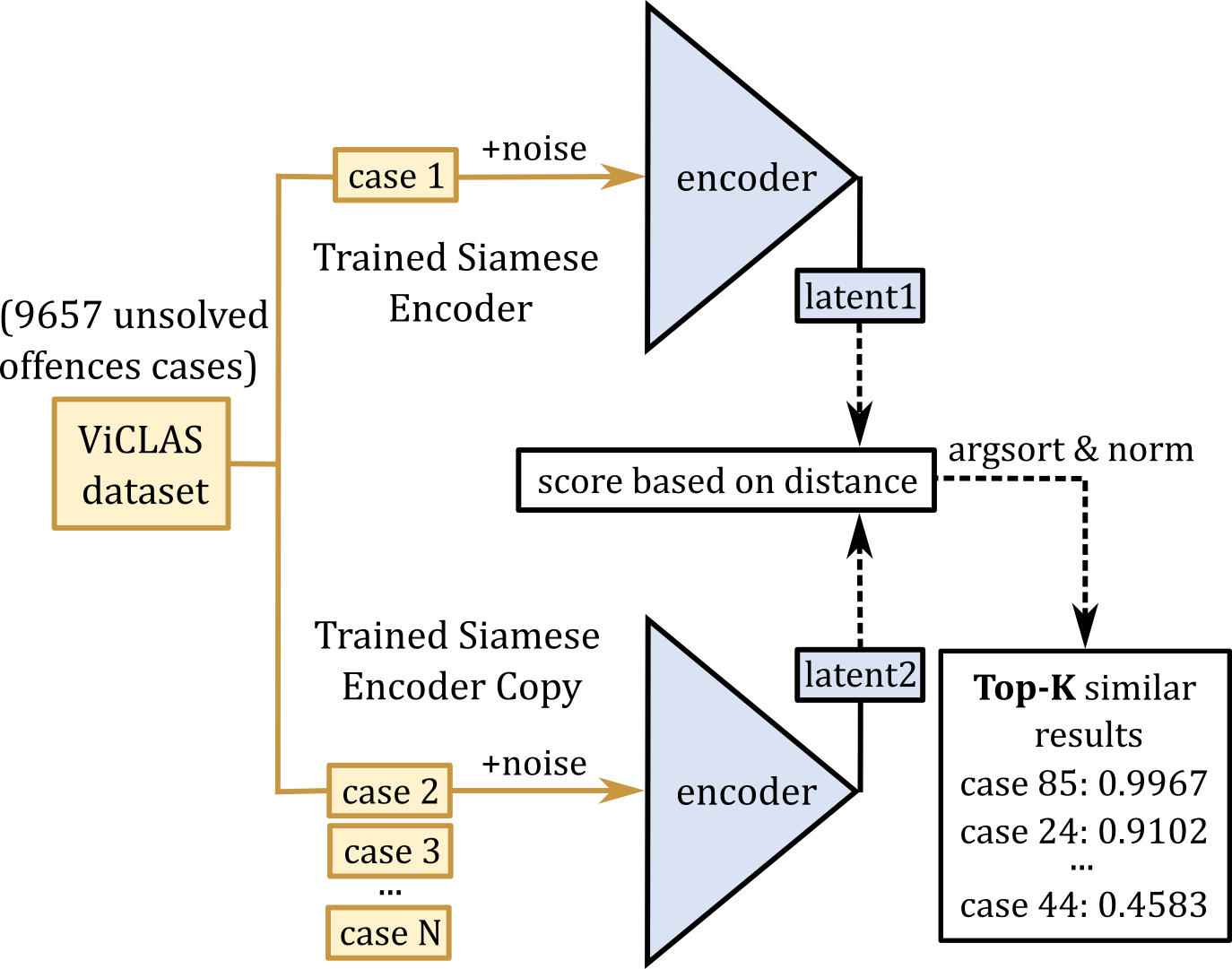}
\caption{Network inference pipeline overview. Two identical Siamese Encoders process case pairs to compute latent representations, rank them by distance, and output Top-K similarity scores from the ViCLAS dataset.}
\label{fig:infer}
\end{figure}
\subsection{Model Inference On Unsolved Cases}
\label{sec:infer}
\refFig{infer} illustrates the model inference process. 
To assess the similarity between cases in probability scores, we utilize the latent code generated by our network's encoder. 
For each pair of cases $i$ and $j$, we compute the Euclidean distance between their latent codes $\mathbf{e}_i$ and $\mathbf{e}_j$:
\begin{equation}
D_{ij} = \|\mathbf{e}_i - \mathbf{e}_j\|_2 = \sqrt{\sum_{k=1}^{d} (e_{ik} - e_{jk})^2},
\end{equation}
where $d$ refers to the dimensionality of the latent code. 
To transform this distance into a probability score $S_{ij}$, we apply an exponential decay function:
\begin{equation}
S_{ij} = \exp(-D_{ij}/\beta),
\end{equation}
where $\beta$ is a scaling parameter set to $m/1.5$ to align with our model's training margin. This transformation maps distances to a bounded similarity score $S_{ij} \in (0, 1]$, with an exponential decay that reflects the learned margin boundary.

\subsection{Data Filtering and Cleaning}
\label{sec:data_cleaning}
Each case in the \ViCLAS dataset is split across multiple Excel sheets, introducing potential inconsistencies and incomplete records. Our data cleaning process involves two distinct merging operations. 
First, we handle cases where a single crime incident (identified by ID) has multiple entries within the same sheet by applying a binary encoding rule: if any part of the incident involves a specific attribute (e.g., weapon use in one part but not others), we encode it as 1 in the merged record. This approach loses information about attribute frequency within cases but preserves behavioral presence indicators crucial for linkage analysis.
Second, we unify categorical labels with overlapping meanings (e.g., merging different naming conventions for the same attribute) and label each binary-encoded dimension as either behavioural, contextual, or both. This isolates specific feature subsets for targeted analysis, such as single offender or single victim scenarios.

\subsection{Data Reduction}
\label{sec:data_mapping}
Our data reduction approach addresses the inherent sparsity and high dimensionality of \ViCLAS data through expert-informed feature consolidation of 446 binary dimensions with approximately 91\% zero values, which creates challenges for pattern learning due to overwhelming inactive features~\cite{lim2021performance}.

\begin{table}[h!]
  \centering
  \footnotesize
  \begin{tabular}{lll}
  \toprule
  Strategy & Features Remaining & Reduction Rate \\
  \midrule
  No Map   & 446 (original)     & 0\%          \\
  Map 1    & 282               & 36.8\%       \\
  Map 2    & 384               & 13.9\%       \\
  Map 3    & 266               & 40.4\%       \\
  Map 4    & 217               & 51.3\%       \\
  Map 5    & 286               & 35.9\%       \\
  \bottomrule
  \end{tabular}
  \caption{Comparison of data reduction strategies showing feature count and reduction rates. }
  \label{tbl:feature_mapping}
\end{table}
\textbf{Expert-Driven Mapping Strategy.} As shown in \refTbl{feature_mapping}, We developed five data reduction strategies through collaboration with \NCA domain experts, each capturing different operational perspectives on behavioral crime linkage. This hierarchical framework groups semantically similar variables under abstract categories, identifying behaviorally similar cases despite surface-level differences~\cite{woodhams2007marine}.
Our mapping strategies were developed as follows: \emph{Map 1} was created by an \NCA analyst with 20+ years experience, focusing on investigative relevance and operational utility. \emph{Map 2} refined Map 1 through consultation with forensic psychologists, emphasizing behavioral consistency principles. \emph{Map 3} merged Maps 1 and 2 by selecting more abstract variables where they diverged, creating a hybrid approach. \emph{Map 4} was developed by forensic psychology experts based on crime linkage literature, prioritizing behavioral distinctiveness. \emph{Map 5} represented a refined version of Map 4 with reduced abstraction to maintain behavioral specificity. For details of data reduction strategies, please refer to the Supplementary Materials. 

\begin{figure}[ht!]
  \centering
  \includegraphics[width=0.47\textwidth]{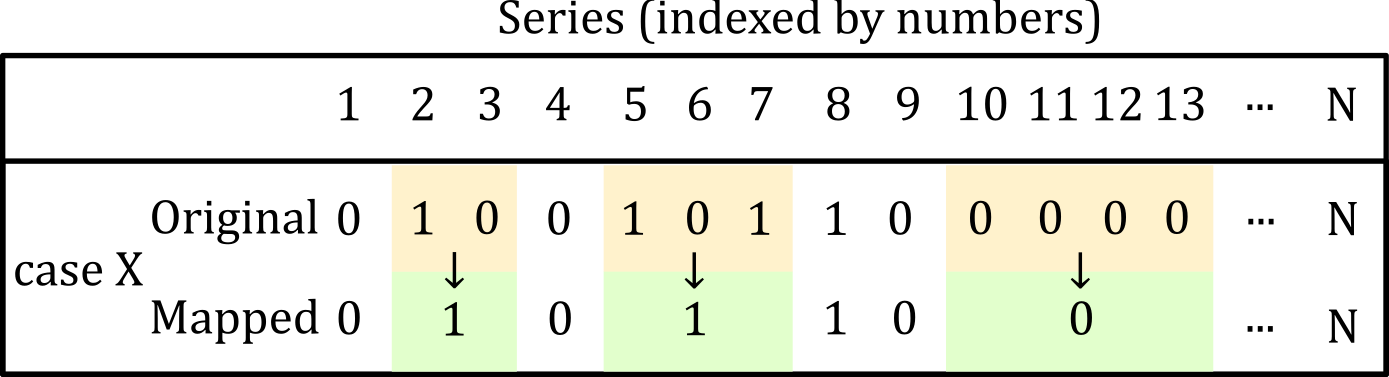}
  \caption{Schematic figure of data reduction. 
  The original binary-coded features are consolidated into broader categories to reduce dimensionality and improve representation.
}
\label{fig:mapping}
\end{figure}
\textbf{Consolidation Methodology.} \refFig{mapping} illustrates our reduction process, where contextually similar features are aggregated into broader categories. For example, location-specific variables such as "shopping mall car park" and "sports complex car park" are consolidated under the abstract category "car park," preserving semantic meaning while reducing dimensionality. This approach aligns with findings that analysts often link crimes thematically rather than through exact behavioral matching~\cite{davies2018practice}.

\textbf{Validation and Selection.} Our evaluation across six configurations (including unmapped data) demonstrates that strategic dimensionality reduction enhances model performance while preserving investigative relevance. Map 5 emerges as optimal, achieving 84\% AUC while reducing features by 44.8\%, suggesting that moderate abstraction preserves behavioral patterns while eliminating noise. See Supplementary Material for detailed mapping specifications and \refTbl{performance_metrics_extend} for comprehensive performance comparisons.

\subsection{Implementation}
We implement our approach using PyTorch~\cite{pytorch}. 
The training was conducted on an Intel 12700k CPU.
The model was trained for 2 epochs with a batch size of 128, the total training time was 6 hours.
We utilized the Adam optimizer~\cite{kingma2014adam} with hyperparameters $\beta_1 = 0.9$, $\beta_2 = 0.999$, the learning rate of $0.001$, and the Cosine Annealing Learning Rate~\cite{CosineAnnealingLR} as scheduler. 
Data augmentation included the addition of Gaussian noise to inputs to avoid gradient vanishing issues.
The dataset was partitioned using 5-fold \CV for model training and evaluation. For more information about data leakage related issues due to \CV, please refer to the Supplementary Materials.

\begin{table}[ht]
  \centering
  \footnotesize
  \setlength{\tabcolsep}{2pt}
  \begin{tabular*}{\columnwidth}{@{\extracolsep{\fill}} l c c c c c }
  \hline
  \multirow{2}{*}{\textbf{Method}} &
  \multicolumn{2}{c}{\textbf{AUC}} & \multicolumn{2}{c}{\textbf{TP FP}} & \textbf{AUPRC} \\
  \cline{2-6}
  & Mean & Std & Mean & Std & Mean \\
  \hline
  Ours                   & 85 & 1.98 & \textbf{77.73} & 4.32 & -- \\
  Logistic Regression    & \textbf{86} & 2.14 & 77.19          & 5.98 & -- \\
  PCA                   & 82 & 4.02 & 64.97          & 4.00 & -- \\
  \hline
  \end{tabular*}
  \caption{Summary of performance metrics on the Single Victim-Offender-Scene Series dataset. AUPRC is not reported for this dataset.  \textit{PCA}: Principal Component Analysis.}
  \label{tbl:performance_metrics_init}
\end{table}

\begin{table}[ht]
  \centering
  \footnotesize
  \setlength{\tabcolsep}{1.5pt}
  \begin{tabular*}{\columnwidth}{@{\extracolsep{\fill}} l c c c c c }
  \hline
  \multirow{2}{*}{\textbf{Method}} &
  \multicolumn{2}{c}{\textbf{AUC}} & \multicolumn{2}{c}{\textbf{TP FP}} & \textbf{AUPRC} \\
  \cline{2-6}
  & Mean & Std & Mean & Std & Mean \\
  \hline
  Ours                     & 77 & 2.11 & 68.31 & 1.92 & 13.32 \\
  Ours (map 1)             & 83 & 2.34 & 78.21 & 2.34 & 15.17 \\
  Ours (map 2)             & 77 & 2.29 & 69.12 & 2.45 & 13.97 \\
  Ours (map 3)             & 80 & 3.43 & 76.53 & 2.71 & 14.04 \\
  Ours (map 4)             & 76 & 3.01 & 74.04 & 3.12 & 14.51 \\
  Ours (map 5)             & \textbf{84} & 2.86 & \textbf{79.38} & 2.56 & \textbf{15.43} \\
  Logistic Regression      & 75 & 2.97 & 70.43          & 2.12 & 10.24 \\
  Naive Siamese            & 76 & 2.15 & 67.53          & 2.60 & 13.45 \\
  Naive Siamese (map 1)    & 81 & 2.55 & 75.12          & 2.44 & 14.56 \\
  Naive Siamese (map 2)    & 74 & 2.97 & 68.36          & 2.57 & 13.41 \\
  Naive Siamese (map 3)    & 79 & 3.34 & 74.47          & 2.85 & 13.48 \\
  Naive Siamese (map 4)    & 74 & 3.19 & 72.08          & 3.28 & 14.12 \\
  Naive Siamese (map 5)    & 83 & 2.72 & 76.20          & 2.69 & 15.09 \\
  \hline
  \end{tabular*}
  \caption{Summary of performance metrics on the Multiple Victim-Offender-Scene Series dataset. \textit{Naive Siamese} refers to the basic Siamese network from~\cite{solomon2020crime}.}
  \label{tbl:performance_metrics_extend}
\end{table}

\section{Evaluation} 
In this section, we systematically evaluate our Siamese Autoencoder approach for crime linkage analysis by addressing five core \RQ (\RQ1 to \RQ5). 
Each subsection focuses on one research question, detailing the corresponding experiments, metrics, and findings to clarify the purpose and outcome of our solution.

\subsection{\RQ\!1: Effectiveness Compared to Baselines}
To address \textbf{RQ1}, we evaluate our method using \AUC, \TPFP, and \AUPRC metrics across 5-fold validations while introducing domain-specific data reduction strategies (See Supplementary for AUPRC details). On the initial dataset (\refTbl{performance_metrics_init}), our method performs comparably to logistic regression in \AUC and TP Fixed FP and outperforms \PCA, though logistic regression achieves better overall performance due to the limited dataset size, evidenced by its weaker performance on the extended dataset.

On the extended dataset (\refTbl{performance_metrics_extend}), we implement various data reduction strategies with map 5 proving most effective. Our approach outperforms both logistic regression and Naive Siamese~\cite{solomon2020crime}, which uses standard twin networks with shared weights, concatenating geo\&temp data with network input rather than adding it individually in the decoder stage.  
With map 5, our approach achieves 84\% \AUC, 79.38\% TP Fixed FP, and 15.43\% \AUPRC, representing relative improvements over Logistic Regression (12.0\%, 12.71\%, and 50.68\% respectively) and Naive Siamese (6.67\%, 8.37\%, and 9.36\% respectively), calculated as  \((\text{Ours} - \text{Baseline}) / \text{Baseline}\). These improvements are robust across all metrics, exceeding one standard deviation in multiple configurations (Map 5: 84\% ± 3\% AUC vs. 75\% ± 3\% for logistic regression), with the 51\% AUPRC improvement proving  practical significance for investigative contexts.

\subsection{\RQ\!2: Impact of Domain-Specific Data Reduction}
Building on the results from \textbf{RQ1}, we conduct a detailed analysis of how different domain-specific data reduction strategies affect model performance. 
Our approach reveals significant variations among different reduction strategies. 
Across the five reduction strategies, \AUC exhibits a range of 8.00 and a \Std of 3.54, TP Fixed FP shows a range of 10.26 and a \Std of 4.07, and \AUPRC varies with a smaller range of 1.46 and a \Std of 0.66.
On average, our method demonstrates an improvement of 2.30\%, 3.02\%, and 3.48\% in AUC, TP Fixed FP, and AUPRC compared to the Naive Siamese baseline. These variations suggest that domain-expert designed data reduction strategies can help maintain semantic relationships and reduce sparsity, thereby improving the model's capacity to identify crime patterns.

\subsection{\RQ\!3: Impact of Architectural and Training Choices}
For \textbf{RQ3}, we benchmark our MLP architecture against two established paradigms – 1D \CNN \cite{1DCNNUNET} with convolutional temporal filters and SIREN \cite{sitzmann2020implicit} using periodic activation functions. 
As shown in ~\refTbl{architecture_comparison}, the MLP-based design achieves a mean AUC of 65.30\%, outperforming both 1D \CNN (57.10\%) and SIREN (52.16\%) variants.
Additionally, our analysis shows that omitting skip connections results in a mean AUC of 63.03\%, outperforming configurations with skip connections (56.48\%) by 6.55\%. 
This suggests that direct feature propagation may interfere with the abstraction of subtle crime patterns, supporting the exclusion of skip connections in our design.
Additionally, network depth analysis identifies an optimal configuration of 2 encoder and 2 decoder layers, achieving the highest AUC of 77.29\%. 
Both shallower architectures (e.g., 1+1 with 52.49\%) and deeper ones (e.g., 4+4 with 63.57\%) show degraded performance. 
The performance of 2+2 layers suggests this configuration balances feature abstraction capacity against overfitting risks in sparse data regimes.
These observations could also provide insight into the future variations of our solution.

\subsection{\RQ\!4: Effect of Geographic-Temporal Integration} 
For \textbf{RQ4}, we examine how integrating geographic-temporal information (geo\&temp) affects our model’s ability to link crimes. As shown in \refTbl{architecture_comparison}, 
embedding this data at the \emph{decoder} level generally yields higher \AUC\ values 
than input-level concatenation. For \MLP in particular, decoder-level integration elevates the \AUC\ from 76.43\% to 77.29\%, an absolute increase of 0.86\%. 
A similar trend holds for the 1D~\CNN\ and SIREN models. Specifically, 
1D~\CNN\ with no skip connections improves from 58.45\% to 61.74\% \AUC\ (+3.29\%),  while SIREN rises from 55.19\% to 58.28\% (+3.09\%).
This benefit appears to stem from allowing the model to refine or “gate” geo\&temp data in the context of previously learned embeddings, rather than merging it directly among high dimension data. In the sparse \ViCLAS\ data setting, an early-stage concatenation risks diluting the geo\&temp signals. In contrast, late-stage integration at the decoder enhances latent codes with geographic-temporal cues.

\begin{table}[ht]
  \centering
  \footnotesize
  \begin{tabular}{l c c c c}
  \hline
  \textbf{Layer} & \textbf{Skip} & \textbf{Depth} & \textbf{geo\&temp} & \textbf{AUC (\%)} \\
  \hline
  1D CNN & \ding{52} & 2+2 & Concat & 53.32 \\ 
  1D CNN & \ding{55} & 2+2 & Concat & 58.45 \\ 
  1D CNN & \ding{52} & 2+2 & Decoder & 54.89 \\ 
  1D CNN & \ding{55} & 2+2 & Decoder & 61.74 \\ 

  SIREN & \ding{52} & 2+2 & Concat & 47.78 \\ 
  SIREN & \ding{55} & 2+2 & Concat & 55.19 \\ 
  SIREN & \ding{52} & 2+2 & Decoder & 54.40 \\ 
  SIREN & \ding{55} & 2+2 & Decoder & 58.28 \\ 

  MLP & \ding{52} & 2+2 & Concat & 61.40 \\ 
  MLP & \ding{52} & 2+2 & Decoder & 67.07 \\ 
  MLP & \ding{55} & 1+1 & Decoder & 52.49 \\ 
  MLP & \ding{55} & 3+3 & Decoder & 70.85 \\ 
  MLP & \ding{55} & 4+4 & Decoder & 63.57 \\ 
  MLP & \ding{55} & 2+2 & Concat & 76.43 \\ 
  MLP (Ours) & \ding{55} & 2+2 & Decoder & 77.29 \\ 

  \hline
  \end{tabular}
  \caption{Comparative analysis of different architectural configurations and their impact on model performance without data reduction strategies.
  \textit{Layer} indicates the architecture type, 
  \textit{Skip} denotes the residual connections, 
  \textit{Depth} specifies the encoder-decoder layer configuration, 
  and \textit{geo\&temp} describes the embedding integration approach 
  (Concat: input-level integration, Decoder: decoder-level integration).
  }
  \label{tbl:architecture_comparison}
\end{table}

\subsection{\RQ\!5: Out of Time Distribution Test}

To address \textbf{RQ5}, we assess temporal generalization through an out-of-time test. We conduct experiments on an additional dataset provided by \NCA containing 1,165 solved cases spanning 2021 to 2025, representing crimes that occurred after our main training period (1990-2021). This temporal separation assesses the model's ability to generalize beyond training distribution and highlights the necessity for periodic model retraining in operational deployment to maintain predictive accuracy as crime patterns evolve over time.
Notably, this post-2021 period coincides with COVID-19 aftermath, where significant behavioral shifts have been observed due to social, economic, and environmental changes \cite{Viclas2022MarkLaw,woodhams2024incidence}, presenting a challenging yet realistic temporal generalization test.

Our analysis reveals two distinct optimization strategies: Map 1 achieves highest recall (77.94\%) with 53 of 68 true positives, optimal for maximizing linkage detection; Map 5 minimizes investigative burden with only 97,013 false positives from $\binom{1165}{2} = 678,030$ possible pairs while maintaining 42.65\% recall, representing the most efficient screening approach.
While out-of-time performance is significantly lower than 5-fold \CV results, our approach maintains practical value by functioning as an effective screening system that reduces manual review workload by up to 80\% while preserving over half of genuine criminal connections. This shows the importance of domain-specific performance analysis in highly imbalanced tasks, where standard precision metrics may not reflect practical utility for end-users. For detailed performance comparisons across all mapping strategies, please refer to the supplementary material.

\section{Discussion}
Our experiments demonstrate that combining behavioural and contextual features enhances linkage performance, whereas using either alone yields suboptimal results, underscoring their complementary roles in capturing offender consistency. Unexpectedly, deeper architectures and skip connections—common optimizations—diminish model effectiveness for \CL, by introducing spurious correlations in binary-encoded behavioural data. Future work should investigate the incorporation of natural‐language offence descriptions to recover complex patterns lost in binary encoding. 
Our out-of-time distribution analysis reveals obvious performance degradation on post-2021 data, coinciding with COVID-19 aftermath where behavioral shifts have been observed. This necessitates periodic model retraining in operational deployment to maintain predictive accuracy as crime patterns evolve.

While our mapping was designed with UK expert knowledge without claiming generalizability beyond UK's \ViCLAS, similar behavioral data systems exist globally across Europe, Canada, and New Zealand~\cite{rcmp_viclas}. Our work demonstrates Siamese networks' potential for behavioral analysis across datasets, building on established precedent of \ML applications to various crime types including burglary and robbery~\cite{tonkin2019linking}.

\textbf{Demographic Representativeness Assessment.} We assessed whether crimes recorded in \ViCLAS were representative of those reported to the unit by UK police forces. Triage processes include offences in \ViCLAS only when containing sufficient behavioral information, potentially introducing bias if certain victims or suspects are associated with crimes lacking such detail. Comparing victim and suspect demographics (age and profession) between unit reports and \ViCLAS revealed no significant differences for available variables. 
No studies currently assess behavioral differences across victim groups or offender demographics, and victim surveys provide no demographic breakdowns for experiences of stranger rape. This gap represents a significant direction for future interdisciplinary research. 

\section{Conclusion}
\label{sec:discussion_and_conclusion}
We proposed a Siamese Autoencoder framework to predict offence linkages in high-dimensional, sparsely distributed, binary-encoded \ViCLAS data, incorporating geographic and temporal information. Experiments on a real-world dataset provided specifically for this research demonstrate superior performance compared to baseline methods. We also evaluate domain-expert-driven data reduction strategies integrated into our training pipeline and find that such reductions can improve both model performance and efficiency.
Future work will extend our method to additional crime types to assess generalizability across offence categories. We also aim to evaluate our method in audited operational settings, examining both effectiveness and ethical implications, to clarify real-world benefits, risks, and guide informed integration of ML tools into offence triage.

\section{Ethics Statement}
When developing and evaluating \CL algorithms, it is essential to account for potential bias in input data and resulting group disadvantage. Training sets predominantly consist of crime series linked by DNA scene-to-scene hits or criminal justice outcomes, while “apparent one-offs” lack confirmed links. Because behavioural differences may exist between solved and unsolved crimes, models trained primarily on solved series risk degraded performance when applied to under-represented groups in unsolved cases. We therefore mapped potential sources of bias across data preparation, model training, and deployment. Our preliminary assessments found no evidence of demographic or geographic bias in the training data, detailed in the Supplementary Material.

\textbf{Deployment Safeguards and Governance.} Operational use of \CL systems requires explicit risk mitigation: (1) \emph{Human-in-the-Loop}: The system provides ranked similarity lists to support, not replace, investigative decision-making, with analysts required to document their reasoning. (2) \emph{Routine Bias Audits}: Continuous monitoring for demographic and geographic disparities is necessary, even where initial assessments indicate no significant bias, as crime patterns and data collection practices change. (3) \emph{Transparent Evaluation}: System performance, assumptions, and limitations must be made clear to prevent over-reliance on automated outputs. (4) \emph{Continuous Adaptation}: Periodic retraining is required to address temporal distribution shifts, as evidenced by performance degradation on post-2021 data coinciding with societal behavioural changes. These measures align with the National Police Chiefs’ Council Covenant for Using AI in Policing and ensure that the approach supports, rather than substitutes, human investigative expertise.

\textbf{Cross-Jurisdictional Adaptation.} Our binary encoding approach is deliberately language-agnostic—crime reports in any language are coded into structured binary features (e.g., "weapon used: yes/no") before entering our pipeline, providing operational safety, cross-border compatibility, and reduced interpretation variance. However, feature taxonomies require cultural adaptation, as weapon categories, location types, and offense characteristics vary across legal systems. Our mapping methodology provides a replicable framework that local domain experts can adapt to jurisdictional standards. Researchers applying our approach to new jurisdictions should: (1) engage local analysts to develop culturally-appropriate feature consolidations, (2) validate that behavioral consistency and distinctiveness principles hold in their crime context, and (3) assess whether geographic-temporal patterns exhibit similar discriminative properties. Further discussion on mapping strategies and generalizability is provided in the supplementary materials.

\section{Acknowledgments}
This research was partially supported by funding from the National Crime Agency, UK. The authors would also like to thank the analysts from the SCAS unit at the National Crime Agency for their valuable assistance with data preparation and for providing insightful feedback on the research findings.

\bibliography{example_paper}


\newpage
\appendix
\onecolumn

\section{Impact Statement}
\label{supp:impact}

Our paper seeks to enhance public safety through improved crime linkage analysis, while acknowledging the ethical sensitivities and dual-use potential of criminal justice technologies.

\textbf{Positive Impacts.} Our method addresses key challenges in serious crime investigations by: (1) \emph{Investigative efficiency}: Accelerating the identification of potential crime series, enabling more effective allocation of limited law enforcement resources; (2) \emph{Public safety}: Facilitating early detection of serial offenders, thereby reducing the risk of further victimization through timely intervention; (3) \emph{Justice advancement}: Linking previously unconnected cases, offering closure for victims and accountability for offenders.

\textbf{Risk Mitigation and Ethical Considerations.} Key risks are proactively addressed: (1) \emph{Algorithmic bias}: Although no demographic disparities were observed in our bias assessments, ongoing monitoring is essential to ensure fairness; (2) \emph{Over-reliance on automation}: The system functions as a decision-support tool, requiring analyst oversight and documented rationale for investigative decisions; (3) \emph{Privacy and confidentiality}: All research adhered to strict data protection protocols and received appropriate ethical approvals.

\textbf{Bias Analysis.}
Addressing demographic imbalances presented significant challenges. Undersampling would reduce dataset size and compromise model performance, while creating synthetic crime data for underrepresented groups is inappropriate, as it would not reflect actual reported incidents. We focused on transparency in evaluation metrics across demographic groups and ensuring human oversight in decision-making.
Our approach serves as a screening tool generating ranked lists of similar cases to support crime analysts. Final decisions rest with analysts, who must complete detailed reports explaining their reasoning for including or excluding potentially linked cases. We conducted workshops with end-users, senior analysts, and managers to identify bias sources throughout the process, adhering to the National Police Chiefs' Council Covenant for Using AI in Policing.

\section{Network Design Motivation}
\label{supp:Architecture}
In this section, we discuss the limitations of alternative deep learning architectures and designs for crime linkage analysis.

\textbf{Siamese Networks for Sparse Crime Data.} Our dataset presents additional challenges with extreme sparsity (~9\% of cases in series) and high dimensionality. Logistic regression, with its linear boundaries, fails to capture the complex behavioral interactions essential for crime linkage~\cite{berk2013statistical,tollenaar2019optimizing}. In contrast, Siamese networks excel in sparse, high-dimensional scenarios by learning latent representations that capture interaction effects between features~\cite{pei2016siamese} and demonstrate superior performance over logistic regression under limited and highly imbalanced tabular data conditions~\cite{basu2022siamese}—precisely matching our \ViCLAS dataset characteristics.

\textbf{Data Representation and Dense Embeddings.} 
The use of dense embeddings (common in NLP tasks) was considered as an alternative to sparse binary representations. However, confidentiality constraints from the UK's \NCA restrict the usage of detailed textual descriptions or dense latent embeddings derived from sensitive crime details. Binary encoding thus remains standard practice for crime linkage research due to its balance between operational safety, confidentiality, and analytical utility.

\textbf{Transformer.} Although Transformer is proven to be powerful for sequential data, it suffers from fundamental mismatches with crime linkage requirements. As self-attention assumes meaningful relationships between token positions, but crime features represent unordered categorical attributes (e.g., weapon type, location, victim characteristics) with no inherent sequential structure~\cite{tay2020efficient}. Positional encoding also becomes meaningless when applied to arbitrarily ordered behavioral categories. The attention mechanism designed to capture long-range dependencies in sequences instead creates spurious correlations between unrelated crime attributes.

\textbf{CNN.} Standard CNNs also apply problematic assumptions for categorical crime data. Convolutional operations assume local spatial relationships and translation invariance—properties absent in behavioral feature vectors where adjacent positions carry no semantic proximity~\cite{lecun1998gradient}. A CNN applied to crime data would treat neighboring features (e.g., "knife used" and "outdoors location") as spatially related when they represent entirely different behavioral dimensions. Additionally, Pooling or Transpose Convolution operations will destroy the precise semantic meaning, which is crucial for behavioral recognition in sparse binary representations.

\section{Detailed Out-of-Time Distribution Analysis}
Table \ref{tbl:out_of_time_results} presents comprehensive performance metrics across six data reduction strategies, revealing distinct operational trade-offs for criminal investigation workflows. Among these approaches, Map 1 demonstrates the strongest linkage detection capability, identifying 53 of 68 true linked pairs (77.94\% recall) using 282 features. While this strategy minimizes missed criminal connections with only 15 false negatives, it necessitates reviewing 399,461 cases, representing the highest investigative burden across all evaluated methods.

In contrast, Map 5 achieves optimal screening efficiency by generating only 97,013 false positives while maintaining 42.65\% recall through 286 features. This approach reduces manual review workload by 75\% compared to Map 1's recall-optimized strategy, successfully detecting 29 true linkages and proving particularly suitable for resource-constrained investigative environments. Between these extremes, Map 4 exhibits the highest overall discriminative ability with an AUC of 71.62\% using 217 features, though this translates to moderate operational performance characterized by 48 true positives and 228,421 false positives.

The baseline raw feature approach, employing all 446 available features, achieves balanced yet suboptimal performance with 51.47\% recall and 174,549 false positives. This baseline demonstrates that strategic feature reduction through domain-specific mapping enhances both recall and efficiency compared to using the complete feature set. Notably, performance does not correlate linearly with feature dimensionality, as evidenced by Map 5's superior efficiency despite using 286 features compared to Map 2's 384 features, and Map 1's enhanced recall performance with 282 features versus the raw approach's 446 features. This indicates that feature quality and domain relevance supersede mere quantity in criminal linkage analysis applications.

\begin{table}[ht]
  \centering
  \footnotesize
  \begin{tabular}{l c c c c c c c}
  \hline
  \textbf{Strategy} & \textbf{AUC} & \textbf{TP@15\%FP} & \textbf{TP} & \textbf{FP} & \textbf{TN} & \textbf{FN} & \textbf{Recall} \\
  \hline
  Raw (446 features)    & 61.06 & 30.88 & 35 & 174,549 & 503,413 & 33 & 51.47 \\
  Map 1 (282 features)  & 61.97 & 26.47 & \textbf{53} & 399,461 & 278,501 & \textbf{15} & \textbf{77.94} \\
  Map 2 (384 features)  & 69.16 & 42.65 & 40 & 179,186 & 498,776 & 28 & 58.82 \\
  Map 3 (266 features)  & 70.20 & 42.65 & 39 & 163,253 & 514,709 & 29 & 57.35 \\
  Map 4 (217 features)  & \textbf{71.62} & 38.24 & 48 & 228,421 & 449,541 & 20 & 70.59 \\
  Map 5 (286 features)  & 65.66 & 42.65 & 29 & \textbf{97,013} & \textbf{580,949} & 39 & 42.65 \\
  \hline
  \end{tabular}
  \caption{Out-of-time distribution test results across all data reduction strategies. TP@15\%FP indicates True Positive Rate at 15\% False Positive Rate. Bold values indicate best performance per metric.}
  \label{tbl:out_of_time_results}
\end{table}

Table \ref{tbl:post2021_statistics} provides detailed statistics for the post-2021 dataset used in RQ5, spanning 2021 to 2025.

\begin{table}[ht]
\centering
\footnotesize
\begin{tabular}{lcccccc}
\hline
\textbf{Category} & \textbf{Solved} & \textbf{Unsolved} & \textbf{Both} & \textbf{Missing} & \textbf{Total} \\
\hline
All       & 1,165 & 972  & 2,137 & 0 & 2,137 \\
One-offs  & 1,049 & 972  & 2,021 & 0 & 2,021 \\
In series & 116   & 0    & 116   & 0 & 116   \\
\hline
\end{tabular}
\caption{Post-2021 dataset statistics (2021--2025). The dataset contains 2,137 total cases, with 94.6\% classified as apparent one-offs and 116 crime series cases providing 68 linkable pairs for temporal generalization testing.}
\label{tbl:post2021_statistics}
\end{table}

\section{Data Reduction Strategies}
\label{supp:data_reduction}

This section provides detailed methodology for our feature consolidation approach, building on established \ViCLAS reliability studies \cite{woodhams2019linking,snook2012violent}. Our systematic approach preserves critical behavioral patterns while addressing the computational challenges posed by high-dimensional sparse data.

\textbf{Methodological Foundation.}
Our reduction strategies were grounded in variable categories demonstrating strong inter-rater agreement from prior reliability studies. Following \cite{woodhams2019linking}, we categorized variables into behavioral features (offender actions including approach methods, control mechanisms, and forensic awareness measures) and contextual features (environmental and situational aspects such as location type and temporal patterns). This categorization informed our consolidation decisions, prioritizing variables with demonstrated investigative utility and behavioral consistency.

\textbf{Consolidation Methodology.}
We implemented hierarchical grouping of semantically related variables under abstract categories. To illustrate this approach, consider weapon-related variables, identified by \cite{snook2012violent} as reliability-challenging. Original encoding included 16 dimensions spanning specific weapon types, acquisition patterns, and usage contexts. Our consolidation reduced these to four key dimensions: general firearm presence, edged weapon use, blunt object use, and acquisition strategy (planned vs. opportunistic). This aligns with \cite{snook2012violent}'s finding that weapon class categorization achieves substantially higher inter-rater reliability (Cohen's $\kappa=.62$) compared to specific implement identification ($\kappa=.19-.34$).

\textbf{Strategy Development Process.}
Maps 1 and 2 were developed independently by at least two experts each—Map 1 by \NCA analysts (including one senior analyst with 20+ years experience) and Map 2 by forensic psychologists. For each mapping, experts first conducted independent consolidations, then convened to resolve conflicts through structured discussion with senior analyst oversight. This ensured final mappings reflected consensus rather than individual preferences. Map 3 created a hybrid approach by selecting more abstract variables where Maps 1 and 2 diverged. Maps 4 and 5 implemented literature-driven abstractions developed by forensic psychology experts, with Map 5 achieving optimal performance through reduced abstraction that maintained behavioral specificity while preserving investigative relevance.

Each strategy balances three core linkage principles from \cite{woodhams2019linking}: behavioral consistency through \MO preservation, distinctiveness via emphasis on rare behaviors, and investigative relevance as determined through practitioner consultation. This systematic approach enhanced computational efficiency while maintaining operational interpretability necessary for deployment.

\textbf{Variable Categorization and Treatment.}
The foundation of our strategy lies in the careful categorization of variables into behavioral and contextual features, following \cite{woodhams2019linking}. Behavioral features encompass offender actions such as approach methods (e.g., surprise vs. con approach), control mechanisms (verbal threats, physical restraints), and forensic awareness measures. Contextual features include environmental and situational aspects such as location type and temporal patterns. To illustrate our methodology, consider the treatment of weapon-related variables, which \cite{snook2012violent} identified as particularly challenging for reliability. The original encoding included 16 dimensions spanning firearm types (pistol, rifle, shotgun), edged weapons (knife, machete, scissors), blunt objects, and improvised weapons, along with acquisition and usage patterns. Our reduction condensed these into four key dimensions: general firearm presence, edged weapon use, blunt object use, and weapon acquisition strategy (planned vs. opportunistic). This consolidation aligns with \cite{snook2012violent}'s finding that weapon class categorization (Cohen's kappa coefficient $K=.62$) demonstrates substantially higher reliability than specific implement identification ($K=.19-.34$), where $K$ measures inter-rater agreement with values ranging from -1 to 1, with higher values indicating stronger agreement.

The success of this reduction strategy stems from its alignment with fundamental linkage principles outlined in \cite{woodhams2019linking}: behavioral consistency through \MO preservation, distinctiveness via emphasis on rare behaviors, and maintenance of investigative relevance as determined through practitioner consultation. Each of our five mapping strategies represents a different balance of these principles, with Map 5 ultimately achieving optimal performance through its emphasis on behaviorally stable and investigatively relevant features while maintaining sufficient distinctiveness. By carefully balancing dimensionality reduction with information preservation, we achieved a representation that enhances computational efficiency while maintaining the interpretability necessary for operational deployment.

\section{Data Considerations}
\subsection{Cross-Validation and Data Leakage Prevention}
Given the temporal nature of our dataset (spanning 1990-2021) and the presence of crime series (multiple offences by the same offender), we implemented specific measures to prevent data leakage during \CV that could artificially inflate performance metrics. The most critical aspect of our validation strategy is ensuring that all offences from the same crime series are assigned to the same fold, preventing the model from learning linkage patterns from partial series information during training and then being tested on remaining offences from the same series. While our dataset spans 31 years, we do not use strict temporal splits (e.g., training on early years and testing on later years) because many crime series span multiple years (average series lasting approximately 2,847 days), making temporal splits impractical without breaking series integrity. Instead, our random 5-fold partitioning ensures balanced representation of different time periods across all folds while maintaining series-level integrity. During training, the model processes crime pairs $(O_i, O_j)$ with binary labels indicating whether they belong to the same series, and our fold assignment ensures that if either $O_i$ or $O_j$ appears in the validation fold, the pair is excluded from training. For each fold, we construct validation pairs only from offences within that fold, ensuring complete independence from training data, and the final reported metrics represent the average performance across all five folds.

\subsection{Code and Data Availability}
Due to data-sharing agreements with the UK's National Crime Agency, we cannot release the full code, as it includes sensitive features from the ViCLAS database. However, we will release a sanitized version with simulated data that includes our model architecture, training, and evaluation code upon publication at: \url{https://github.com/AlberTgarY/CrimeLinkageSiamese}.

\textbf{Simulation Protocol.} To enable meaningful pipeline stress-testing, our simulated dataset will preserve key statistical properties of the original ViCLAS data:
(1) \emph{Sparsity}: Binary feature vectors maintain $\sim$91\% zero values, matching the observed sparsity in behavioral-contextual encodings. 
(2) \emph{Dimensionality}: Simulated data spans 217-446 dimensions corresponding to the range of our mapping strategies (\refTbl{feature_mapping}).
(3) \emph{Class Imbalance}: The ratio of linked to unlinked pairs ($\sim$1.8\% positive class) is preserved to replicate the operational challenge.
(4) \emph{Geographic-Temporal Properties}: Continuous spatial and temporal features follow distributions fitted to the original data to maintain realistic inter-crime relationships.
The complete simulation code, including data generation procedures and validation metrics comparing simulated vs. real statistical properties, will be released alongside the model weights (available to researchers with appropriate institutional ethics approvals).

\section{AUPRC Calculation}
\label{supp:auprc}

The Area Under the Precision-Recall Curve (AUPRC) quantifies model performance under severe class imbalance, a critical consideration for crime linkage where linked pairs constitute only 1.8\% of potential comparisons. Unlike ROC-AUC, which becomes overly optimistic in imbalanced scenarios, AUPRC emphasizes precision at varying levels of recall—operationally critical for minimizing investigative overhead while ensuring serial offender detection. For predicted similarity scores $S_{ij}$ between crime pairs $(i,j)$, precision and recall are calculated as $TP/(TP + FP)$ and $TP/(TP + FN)$ respectively, where $TP$, $FP$, and $FN$ denote true positives, false positives, and false negatives. The precision-recall curve is generated by sweeping a similarity threshold $\tau \in [0,1]$, with AUPRC computed via trapezoidal integration over 100 uniformly spaced thresholds.

During 5-fold \CV, similarity scores for all validation pairs are first computed and sorted in descending order. Precision and recall values are then evaluated at each threshold, with final AUPRC scores averaged across folds. This macro-averaging approach ensures equal weighting of each fold's contribution, mitigating variability from localized crime patterns. The trapezoidal integration method approximates the area under the curve as $\sum_{k=1}^{n-1} (R_{k+1} - R_k) \cdot (P_k + P_{k+1})/2$, where $(R_k, P_k)$ represent recall-precision coordinates at threshold $\tau_k$.


\end{document}